\documentclass[journal,final,10pt,twocolumn,a4paper]{IEEEtran}

\usepackage{cite}
\usepackage{graphicx}
\usepackage{booktabs}
\usepackage{amsmath}
\usepackage{amsfonts}
\usepackage{multirow}
\usepackage{url}

\begin{document}

\title{A semi-holographic hyperdimensional representation system for hardware-friendly cognitive computing}

\author{
\IEEEauthorblockN{A. Serb\IEEEauthorrefmark{1}, I. Kobyzev\IEEEauthorrefmark{2}, J. Wang\IEEEauthorrefmark{1}, T. Prodromakis\IEEEauthorrefmark{1}}
\IEEEauthorblockA{\IEEEauthorrefmark{1}Zepler Institute, University of Southampton, SO17 1BJ, UK}
\IEEEauthorblockA{\IEEEauthorrefmark{2} David R. Cheriton School of Computer Science, University of Waterloo, N2L 3G1, Canada \\
Corresponding author email: A.Serb@soton.ac.uk}
}

\maketitle

\begin{abstract}
One of the main, long-term objectives of artificial intelligence is the creation of thinking machines. To that end, substantial effort has been placed into designing cognitive systems; i.e. systems that can manipulate semantic-level information. A substantial part of that effort is oriented towards designing the mathematical machinery underlying cognition in a way that is very efficiently implementable in hardware. In this work we propose a `semi-holographic' representation system that can be implemented in hardware using only multiplexing and addition operations, thus avoiding the need for expensive multiplication. The resulting architecture can be readily constructed by recycling standard microprocessor elements and is capable of performing two key mathematical operations frequently used in cognition, superposition and binding, within a budget of below $6\,pJ$ for 64-bit operands. Our proposed `cognitive processing unit' (CoPU) is intended as just one (albeit crucial) part of much larger cognitive systems where artificial neural networks of all kinds and associative memories work in concord to give rise to intelligence.
\end{abstract}

\IEEEpeerreviewmaketitle

%\nocite{IEEEexample:BSTcontrol}

\section{Introduction}\label{intro}

The explosive scale of research output and investment in the field of artificial intelligence (AI) and machine learning (ML) testify to the tremendous impact of the field to the world. Thus far this has manifested itself as a mass-scale proliferation of artificial neural network-based (ANN) algorithms for data classification. This covers multiple data modalities such as most prominently images \cite{Krizhevsky2012} and speech/sound \cite{Hannun2014}, and relies on a number of standard, popular ANN architectures, most notably multi-layer perceptrons \cite{LeCunn2015}, recurrent NNs (in particular, LSTM \cite{Greff2017} and GRU \cite{Wang2015}) and convolutional NNs \cite{Spoerer2017} amongst many others \cite{Kaplan2001, Schurmann2005}.

Thus far the vast majority of market-relevant ANN-based systems belong to the domain of statistical learning, i.e. perform tasks which can be generally reduced to some sort of pattern recognition and interpolation (in time, space, etc.). This, though demonstrably useful, is akin to memorising every answer to every question plus some ability to cope with uncertainty. In contrast, higher level intelligence must be able to support fluid reasoning and syntactic generalisation, i.e. applying previous knowledge/experience to solve novel problems. This requires the packaging of classified information generated by traditional ANNs into higher level variables (which we may call `semantic objects'), which can then be fluently manipulated at that higher level of abstraction. A number of cognitive architectures have been proposed to perform such post processing, most notably the ACT-R architecture \cite{Anderson1997} and the semantic pointer architecture (SPA) \cite{Eliasmith}, which is an effort to manipulate symbols using neuron-based implementations.

Handling the complex interactions/operations between semantic objects requires both orderly semantic object representations and machinery to carry out useful object manipulation operations. Hyperdimensional vector-based representation systems \cite{Plate1995} have emerged as the de facto standard approach and are employed in both the SPA and ACT-R. Their mathematical machinery typically includes generalised vector addition (combine two vectors in such way that the result is as similar to both operands as possible), vector binding (combine two vectors in such way that the result is as dissimilar to both operands as possible) and normalisation (scale vector elements so that overall vector magnitude remains constant). These operations may be instantiated in holographic (all operands and results have fixed, common length) or non-holographic manners. Non-holographic systems have employed convolution \cite{Schonemann1987} or tensor products \cite{Smolensky1990} as binding. Holographic approaches have used circular convolution \cite{Plate1995} and element-wise XOR \cite{Kanerva1997}. Meanwhile, element-wise addition tends to remain the vector addition operation of choice across the board.

Finally, whichever computational methodology is adopted for cognitive computing must be implementable in hardware with extremely high power efficiency in order to realise its full potential for practical impact. This is the objective pursued by a number of accelerator architectures spanning from limited precision analogue neuron-based circuits \cite{Akopyan2015}, through analogue/digital mixtures \cite{Neckar2019} to fully analogue chips seeking to emulate the diffusive kinetics of real synapses \cite{Qiao2015}. More recently memristor-based architectures has also emerged \cite{Rahimi2018}.

In this work, we summarise an existing, abstract mathematical structure for carrying out semantic object manipulation computations and propose an alternative, hardware-friendly instantiation. Our approach uses vector concatenation and modular addition as its fundamental operations (in contrast to the more typical element-wise vector addition and matrix-vector multiplication respectively). Crucially, the chosen set of operations no longer forms a holographic representation system. This trades away some `expressivity' (ability to form semantic object expressions within limited resources) in exchange for compression: Unlike holographic representations semantic object vector length depends on its information content. Furthermore, the proposed system avoids use of multiplication completely, thus allowing for both fast and efficient processing in hardware (avoiding both expensive multipliers and relatively slow spiking systems). Finally, we illustrate how the proposed system can be easily mapped onto a simple vector processing unit and provide some preliminary, expected performance metrics based on a commercially available 65nm technology.

\section{Mathematical foundations and motivation}\label{bgnd}

Generalising the series of work on models of associative memory, many of them inspired from the world of optics \cite{Plate1995, Casasent1989, Fisher1987, Paek1987, Willshaw1990, J1969, Aerts2006, Smolensky1990, Kanerva1997}, one may inspect the most abstract algebraic formulation of it. All we need is a commutative ring $R$ with a distance metric dist.

In order to give this mathematical machinery sufficient power to describe cognitive tasks, one must initially specify the ring operations and impose some restrictions on them. The primary operation (addition, denoted by $+$), enables superposition\footnote{In the literature this is typically called `chunking', but this term by itself does not allude strongly enough to the desired simultaneous similarity between operands and result.}; that is the combination of two elements in such way that the result is equidistant from its operands under the metric dist (i.e. for $a, b \in R$, one has $\text{dist}(a+b,a)\approx \text{dist}(a+b,b)$). The secondary operation (multiplication, denoted by $ \ast$) enables binding; that is the combination of two elements in such manner that the result is ideally completely different from both operands. Next, one needs to store a (finite) set of elements of $R$ including both invertible elements, which we call `pointers' (or `roles'), and not necessarily invertible ones, which we call `fillers'. 

Let us now give an example of how such mathematical machinery may give rise to simple cognition. Assume that we have a ring $R$ with the distance dist and the operations, satisfying the desired properties. Also assume that we fixed five elements of $R$: ``$obj$" and ``$col$" are invertible, ``$red$", ``$green$" and ``$car$" are any elements. We can now construct a new element: $s = obj\ast car + col\ast red$, which can be interpreted as a semantic object ``red car". Now one can ask: what colour is this car? The answer can be accessed by performing an algebraic operation: $col^{-1} \ast s = red + col^{-1}\ast obj\ast car$. Then if the term $col^{-1}\ast obj\ast car$ is either close to zero or in some other way does not interfere with the computation of $\text{dist}$, the stored memory element closest to the result of the query is $red$. Mathematically, the query is $argmin_{r\in R} (\text{dist}(col^{-1} \ast s, r)) = red $. Thus, we observe that the mathematical foundation of AI is underpinned by a solid computational/information processing foundation whose functionality must be preserved in any proposed alternative representation system, even if not necessarily via a distance-equipped commutative ring.

The classical realisation of the commutative ring-based cognition principle is the holographic-like memory \cite{Plate1995}. In this case $R$ is defined as follows: the set is a collection of $n$-dimensional real vectors (n-vectors) $\mathbb{R}^n$. The ring operations are the element-wise addition and circular convolution. The distance metric is the simple Euclidean. To define a pointer or a filler one just needs to independently sample each entry of the vector from the normal distribution $\mathcal{N}(0,1/n)$. 

Finally, the operations of the system must be ideally implementable in hardware in a way that minimises power and area requirements. In practice this means that the fundamental superposition and binding operations must rely on energetically cheap building block operations such as thresholding (an inverter), shifts (flip-flop chain), addition (sum of currents on a wire or digital adder) or possibly analogue multiplication (memristor + switch) \cite{Hu2016}. Implementation details will ultimately determine the actual cost of each operation. The main approaches so far either use too many multiply-accumulate (MAC) operations (circular convolution-based binding from \cite{Plate1995} requires $\approx n^2$ MACs/binding), or are applicable only to binary vectors (radix $k=2$) \cite{Kanerva1997}.

\section{Proposed semi-holographic representation system}\label{propHRR}

In this section we provide an intuitive overview followed by a rigorous mathematical explanation of the proposed architecture interwoven with pointers on how our design decisions aim towards hardware efficiency.  Overall, in order to achieve a more hardware-friendly cognitive algebra realisation we trade away some of the mathematical simplicity from the previous section for implementability. The algebraic structure we are using for cognition is no longer a ring, but a rather exotic construction. It consists of an underlying set and two binary operations (superposition, binding).

\subsection{Building a set of semantic objects}

\begin{figure}
\centering
\includegraphics[width=7.5cm]{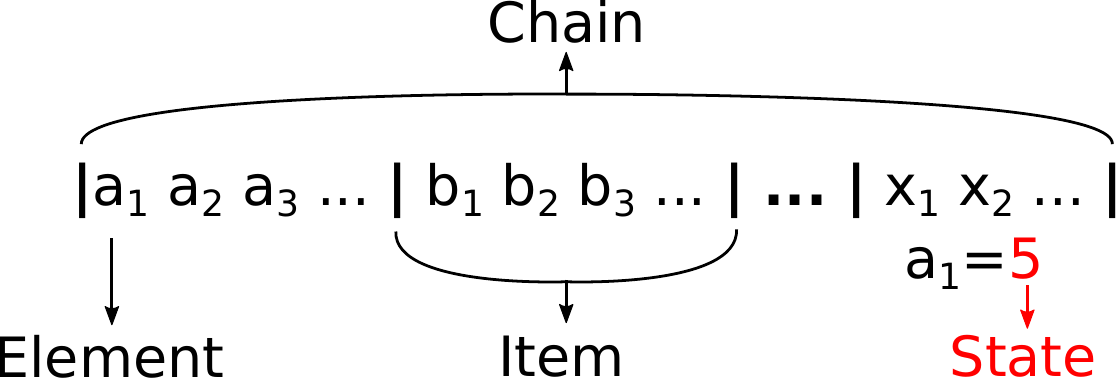}
\caption{Summary of key terms used throughout this text.}
\label{OpSum}
\end{figure}

In our proposed system, the set of semantic objects is perhaps best understood in terms of two subsets: i) Fixed-length `base items', each consisting of $y$ integer elements in the range $[0, p-1]$. The choices of $p$ and $y$ link to desired memory capacity, i.e. the number of semantic objects the system is capable of representing reliably - see section \ref{capacity}). ii) Variable-length `item chains' consisting of multiple concatenated base elements. The maximum length for chains  is $d$ base items for a total of $n$ numerical elements, where $d$ is determined by the hardware design\footnote{However, note that much akin to standard computers being able to process numbers more than 32 or 64 bits, there is no reason why chains longer than $d$ base items cannot be processed using similar techniques.} and affects the capacity of the system to hold/express multiple basic items at the same time. The number of base items in a chain is defined as the rank of the chain. The terminology is summarised in figure \ref{OpSum}

Some observations about our implementation: i) Base items are generally intended for encoding the fundamental vocabulary items of the system (e.g. `red', `apple', `colour') and possible bindings, including the classical `pointer-filler' pairings (e.g. `colour'$\ast$`red': the value of the colour `attribute' is `red'). In contrast, chains are intended for simultaneously holding (superpositions of) multiple base items in memory (e.g. composite descriptions of objects such as: $colour\ast red + object \ast apple$  (`a red apple'), or collections of unrelated items such as: $shape \ast circle + shape \ast square$ (`a circle and a square'). The order in which the superposed items are kept in memory does not bear any functional significance; for the purposes of our system items are either present or absent from a chain. Cognitive systems that are order- or even position-dependent can be, of course, conceived; all that is necessary is for each item to have some mechanism (e.g. a position indicator) for marking its location within a chain. ii) Setting $p, y, d, n$ as powers of 2 offers the attribute of naturally advantageous implementation in digital hardware. This is the approach we choose in this work, as shown in table \ref{objrel}. The choice of $p$ is not necessarily obvious as what constitutes a `good' choice of $p$ will depend on the specific implementations of superposition and binding. iii) Any chain can be zero-padded until it forms a maximum-length chain.

Mathematically the above can be described as follows: Fix natural numbers $p$ and $y$ as above. Then the set of base items is a group $B = (\mathbb{Z}/p)^y$ (under element-wise mod $p$ summation).  The way to form item chains is by executing a direct product of copies of $B$. Then we say that any element of $B^r= \prod_{i=1}^{r}B$ has rank $r$.  The chain of maximal length will be an element of $B^n$, and $n = d \cdot y$.

\begin{table}[t]
\centering
\caption{Summary of relations between basic mathematical objects used in this work. As an example of how to read the table, the top left entry states that: "In every element there are $p=2^l$ states". The term `chain' refers to a maximum size chain. The term 'item' refers to base items. All parameters are integers.}
\begin{tabular}{ll|lll}
                                        &          & \multicolumn{3}{c}{\textbf{In every...}} \\
                                        &          & Element   & Item     & Chain    \\ \hline
\multirow{3}{1.5cm}{\textbf{There are this many...}} & States  & $p=2^l$   & $2^{l+z}$      & $2^{l+z+m}$      \\
                                        & Elements & 1         & $y=2^z$  & $n=2^{z+m}$  \\
                                        & Items    & N/A       & 1        & $d=2^m$ 
\end{tabular}
\label{objrel}
\end{table}

\subsection{Superposition and binding}

Next, we define our set of basic operations. The superposition operation `$+$' is defined as follows: If $a$ and $b$ are semantic objects, then:
\begin{equation}\label{supeq}
	a+b = (a,b)
\end{equation}	
which is a standard direct sum. The result contains both $a$ and $b$ operands preserved completely intact. This can be contrasted with superposition implemented as regular element-wise summation, where each operand is `blurred' and merged into the result. Superpositions of semantic objects whose combined ranks exceed $d$ are not allowed\footnote{In a practical hardware implementation we would either: i) raise an exception and forbid the operation, ii) truncate the result to size $d$ and raise a warning flag or iii) raise a flag and trigger a software sequence (program) designed to handle overlength chains - equivalent to branching to a different subroutine in Assembly language.}. 

Formally speaking, given $a \in B^{d_1},$ $b \in B^{d_2} $, the superposition $a+b$ is just an element $(a,b)$ in a direct products of the  groups $B^{d_1+d_2}$. If $d_1+d_2 > n$, the operation is not defined. 

Next, the binding operation `$\ast$' is defined as a variant of a tensor product between semantic objects where the individual pairings are subsequently subjected to element-wise addition modulo $p$. Mathematically, for given natural numbers $d_1$ and $d_2$, such that $d_1 \cdot d_2 < n$, one can define the binding operation $*: B^{d_1} \times B^{d_2} \to B^{d_1 \cdot d_2}$ by the formula:
\begin{equation}\label{bindeq}
\begin{split}
 (a_1, a_2, \dots, a_{d_1})*(b_1, b_2, \dots, b_{d_2}) = \\
 (a_1 + b_1, a_2 + b_1, \dots, a_{d_1} + b_1, a_1+ b_2, \dots, a_{d_1}+ b_{d_2}) \\
\end{split}
\end{equation}
where $+$ is the group operation in $B=(\mathbb{Z}/p)^y$. One can see that any element from $B$ (base item) is invertible under the binding\footnote{This is where the consequences of our choice of $p$ become apparent: Consider the item consisting of all elements equal to $p/2$. Binding this item to itself twice results in the original item. This becomes problematic if we wish to define a sequence of items as a succession of bindings, e.g. if we define the semantic object `$2$' as $1 \ast 1$, `$3$' as $2 \ast 1$ etc. If, on the other hand $p$ is prime, then for any integer $x \neq 0$ there is a guarantee that if $(k \cdot x) \,\text{mod}\, p = x$, the next greatest solution after $k=1$ is $k=p+1$; this may allow the construction of longer, non-tautological self-bindings vs. non-prime $p$ systems. Morale: the choice of $p$ is not always obvious.}.  

One should notice that modular addition is losslessly reversible: we may indefinitely add and subtract n-vectors, and therefore can perfectly extract any individual term from any multi-term binding combination if we bind with the modulo $p$ summation inverses of all other terms. We also remark that within the context of the order-independence property any binding of chains with length greater than 1 item is effectively a convenient shorthand for describing multiple base item bindings and adds no further computational (or indeed semantic) value.

%\begin{figure}
%\centering
%\includegraphics[width=8.5cm]{Ops_req.pdf}
%\caption{A key cornerstone of the proposed system is using multiple easily hardware-implementable lower level functions to achieve the higher level functionality of superposition and binding. The diagram shows how this arrangement is set up in the present work. Note how only the binding operation changes semantic objects at an element-level.}
%\label{opsumfig}
%\end{figure}

We conclude this section by highlighting that our superposition operation is not length-preserving but our binding is when one of the operands consists of 1 basic item. Thus we describe our system as semi-holographic. Interestingly, this is the opposite of the classical convolution-based system from \cite{Schonemann1987}, where the binding operation is not length-preserving but superposition (element-wise average) is.

\subsection{Similarity metric}

Let us define a distance. First, we  use a ``circular distance" on $\mathbb{Z}/p$: for $a \in \mathbb{Z}/p $, one has $\text{dist}_\circ(a,0) = min(|a|, |p-a|)$, here we also denoted by $a = a + 0\cdot p$ the corresponding representative in $\mathbb{Z}$. For example, for $4 \in \mathbb{Z}/5$, \text{dist}(4,0) = 1. Analogously one defines a distance $\text{dist}_\circ(a,b)$ for any $a, b \in \mathbb{Z}/p $ as $min(|b-a|, p-|b-a|)$. For two vectors $a, b \in B$, one defines the distance as:
\begin{equation}\label{distmet}
    \text{dist}_B(a,b) = \sum_{i=1}^{y}\text{dist}_\circ(a_i,b_i)
\end{equation}

For $a \in B$ and $b = (b_1, \dots b_r) \in B^r$ we define:
\begin{equation}\label{distmet1}
    \text{dist}(a,b) = \text{min}_i\text{dist}_B(a,b_i)
\end{equation}

One can note that  $\text{dist}(a+b,a) = 0$ for any $a \in B$. 

\subsection{Basic properties}

In terms of fundamental mathematical properties: The superposition operation is not closed in general, but it acts as closed when our restriction on the sum of the ranks of the operands is met. It is associative but not commutative. It has an identity element (the empty string), but no inverse operation as such. 

The binding operation is not closed, but acts as closed when the restriction on the product of the ranks of the operands is met. This is always the case when one of the operands is a basic item, i.e. $a \in B$. If $a$ is a basic item, then for any $b \in B^{d}$, we have the commutativity: $a*b = b*a$. If $ a \in B^{d_1}, b \in B^{d_2}, c \in B^{d_3}$, and at least one of $d_i = 1$, then we have associativity: $(a*b)*c = a*(b*c)$. In general it is neither associative nor commutative, however, modulo permutation group on basic item components, it has those properties. 

Finally, one has a distributivity in case of a basic item: $a \in B, b \in B^{d_1}, c \in B^{d_2}$, then $a*(b+c) = a*b+a*c $. In general, as above this property no longer holds (unless we don't care about the order of terms and factorise by the action of permutation group). 

The identity element is the zero element of $B$. All basic elements are invertible under binding.

These properties form a good start for building a cognitive system.

\section{Capacity}\label{capacity}

In terms of higher level properties, a key metric is memory capacity: the maximum number of basic elements storable given some minimum upper bound for memory recall reliability. Each rank 1 semantic object (base item), the smallest type of independent semantic objects, must be uniquely identifiable. As a result, there can be no more than $Q = p^y$ basic memories in total without guaranteeing at least one ambiguous recall, i.e. $Q$ is the maximum memory capacity\footnote{It is very expedient if any semantic object that needs to be stored for quick recall is constructed as a basic object, not in the least because binding any operand with a basic object does not lengthen the operand. For that reason we only consider basic elements when computing memory capacity.}. However, an additional sparsity requirement is necessary in order to guarantee that the system is capable of unambiguously answering queries. Returning to the example from section \ref{bgnd}, in order for the term $col^{-1} \ast obj \ast car$ to be culled from any semantic pointer or filler from our vocabulary it should not coincide with a valid object from the fixed fundamental vocabulary. In order to achieve that, we may impose that our memory safely stores only up to $Q_s$ vocabulary objects, where $s \in \mathbb{R}$ is the desired sparsity factor, and the following formula holds:
\begin{equation}\label{memcap}
    Q_s = \frac{Q}{s}.
\end{equation}

A lower bound for $s$ is given by calculating the number of basic items $J$ that the system can generate given a set of $Q_s$ vocabulary items and allowed complexity. These will all need to be accommodated unambiguously for guaranteeing reliable recall. In our proposed system the only operation that can generate basic items from combinations of vocabulary items is the binding operation. Therefore for $Q_s$ vocabulary items we obtain $ \frac{Q_s^2}{2}$ derived items arising from all the possible unordered (to account for the commutativity) pairwise bindings. This rises to $\frac{Q_s^\gamma}{\gamma!}$ for exactly $\gamma$ allowed bindings, and in general the system can generate:

\begin{equation}\label{sfac}
    J = \sum_{i=0}^\Gamma \frac{Q_s^i}{i!} \approx \frac{Q_s^\Gamma}{\Gamma!} \text{, for } \frac{Q_s}{\Gamma} \gg 1
\end{equation}
basic items, if we allow anything between 0 and $\Gamma$ bindings in total. Ideally we want to account for all possible basic items from the fundamental vocabulary via bindings, so $J = Q (=p^y)$, and therefore we can transform equation \ref{sfac} into:

\begin{equation}\label{totcap}
    Q_s \approx \sqrt[\leftroot{-2}\uproot{2}\Gamma]{\Gamma!} \cdot p^{\frac{y}{\Gamma}}
\end{equation}
revealing how expressivity is traded against capacity, at least in the absence of any further allowances to combat possible uncertainty in the encoding, decoding or recall of semantic objects. Whether this boundary can be reached in practice requires further study as the particular encodings of each basic item will determine whether specific bindings coincide with pre-learnt vocabulary or other bindings. Let us observe that the more binding is allowed in the system, the less fundamental vocabulary it can memorize (hint: $\lim_{x\xrightarrow{}\infty}\sqrt[\leftroot{-2}\uproot{2}x]{x!} = \frac{x}{e}$). This is an example of a trade-off between capacity and complexity. 

Example: if we choose $p=16, y = 128$ and we allow the system to have at most $\Gamma = 20$ bindings, then the upper bound on the length of the core dictionary we can encode is 422 million items.

\section{Additional semantic object manipulations}

In order to complete the description of the proposed system we need to cover three further issues: i) How does the system cope with uncertainty? ii) Since the system is semi-holographic how does the system map multi-item chains to single base items when necessary? In this work we provide some cursory answers as these questions merit substantially deeper study in the own right.

\textit{Dealing with uncertainty:} The implementation of de-noising will strongly depend on the form of the uncertainty present in the system. We may define uncertainty as a probability distribution that encodes how likely it is to obtain semantic object $x'$ when in fact the ground truth is $x$. For example, if the probability density only depends on the `circular distance' (eq. \ref{distmet}) between the $x$ and $x'$ objects\footnote{This alludes to the radial basis functions (RBFs) used in radial basis neurons \cite{Park1991}.} We may use an adaptation of element-wise average for de-noising. The average is computed as the mid-point along the geodesic. In particular, for $a \in \mathbb{Z}/p$ let also denote by the same symbols $a = a + 0\cdot p$ its representative in $\mathbb{Z}$. Also denote by $\Delta = \text{ceil}(\frac{\text{dist}_\circ(a,b)}{2})$. For $a, b \in \mathbb{Z}/p$, if $|b-a| \leq p - |b-a|$, pick a smaller representative with respect to the standard ordering on $\mathbb{Z}$ (say, it is $a$), then $\text{avg}(a,b) = a + \Delta$. If the alternative inequality happens, pick the greater representative (say, it is $b$), then  $\text{avg}(a,b) = (b + \Delta) \ \text{mod} \ p$. In general, for items $a, b \in B$, we  define the average as the element-wise average. 

To this we add the following observations: i) The purpose of the de-noising average is to reconcile multiple, corrupted versions of a single semantic object vector, not combine different vectors into new semantic objects (i.e. $a_i$ is expected to be reasonably close to $b_i$ most of the time). Nevertheless, when used with radically different semantic objects as inputs, it is inescapable to observe that the operation acts very similarly to binding. The effects of using a binding-like operation for denoising (a task usually handled by superposition) are an interesting subject for further study. ii) Different uncertainty descriptors (probability distribution functions) may lend themselves to different de-noising strategies. So will different metrics. iii) Even with fixed underlying probability distribution assumptions, de-noising may be carried out using multiple alternative strategies. Examples applicable to our assumptions would be majority voting (select element-wise mode instead of mean - works best for large number of input sample terms) or median selection.

\textit{Compressing long chains into basic items:} Ideally any cognitive system should be able to take any expression and collapse it into a new memory that can be stored, recalled and used with the facileness that basic items enjoy. In our case this requires compressing chains into the size of a basic item. In principle, any compression algorithm will suffice. Examples could be applying genetic algorithm-like methods \cite{Deb2002} on the items of a chain or combining said items using any multiplication (e.g., circular convolution etc).

We conclude by remarking that the operation of creating a new semantic object can be reasonably expected to be executed orders of magnitude less frequently than any of the other operations. As such, it is possible to dedicate hardware that is both more complex (luxury of using relatively heavy computation) and more remotely located from the core of the semantic object processor (luxury of preventing the layout footprint of the semantic object generator from impacting the layout efficiency of the processor core).

%\section{Application example: Raven's matrix solving}\label{simres}

%-Show Raven's matrix solution in symbolic form.
%-Highlight operations that need to be undertaken.
%-Show it working in hardware.
%-Assess performance.

% ...statement on why the properties we get are enough to build a cognitive system... Suspicion: we won't need decomposition or anything more than rank-1 bindings. Even the deduplication is doubtful. Explain how not having to implement these functionalities greatly simplifies design.

%-Count the number of parallel lines needed.

%NOTE: This section should be titled after the application it demonstrates. Need to think of something useful and pertinent to a 'real world' application (see examples from other Nature group papers). It should be about 50\% an explanation of a specific instantiated system and the rest for simulated behaviour in an application. Stick to about 700-800 words. This leaves approx. 500 for discussion.

%-Show specific example of a full system, explaining all components and defining why they were chosen, e.g. shift-reg, vs. memory or full ALU implementations; choose one and explain why.
%-Examples of bindings and unbindings. Esp. for probabilistic binding case.

\section{Hardware implementation}\label{hardimp}

In this section we examine how the mathematical machinery can be mapped onto a hardware module which we call the `Cognitive Processing Unit' (CoPU). The system receives chains as input operands and generates new chains at its output after executing the requested superposition and/or binding operations. The CoPU is based on a common block-level design blueprint which can then be instantiated as specific CoPU designs. It is at the point of instantiating a particular CoPU design that the values of key parameters $p, y, d$ are decided upon.

\subsection{Hardware system design}

\begin{figure*}
\centering
\includegraphics[width=18cm]{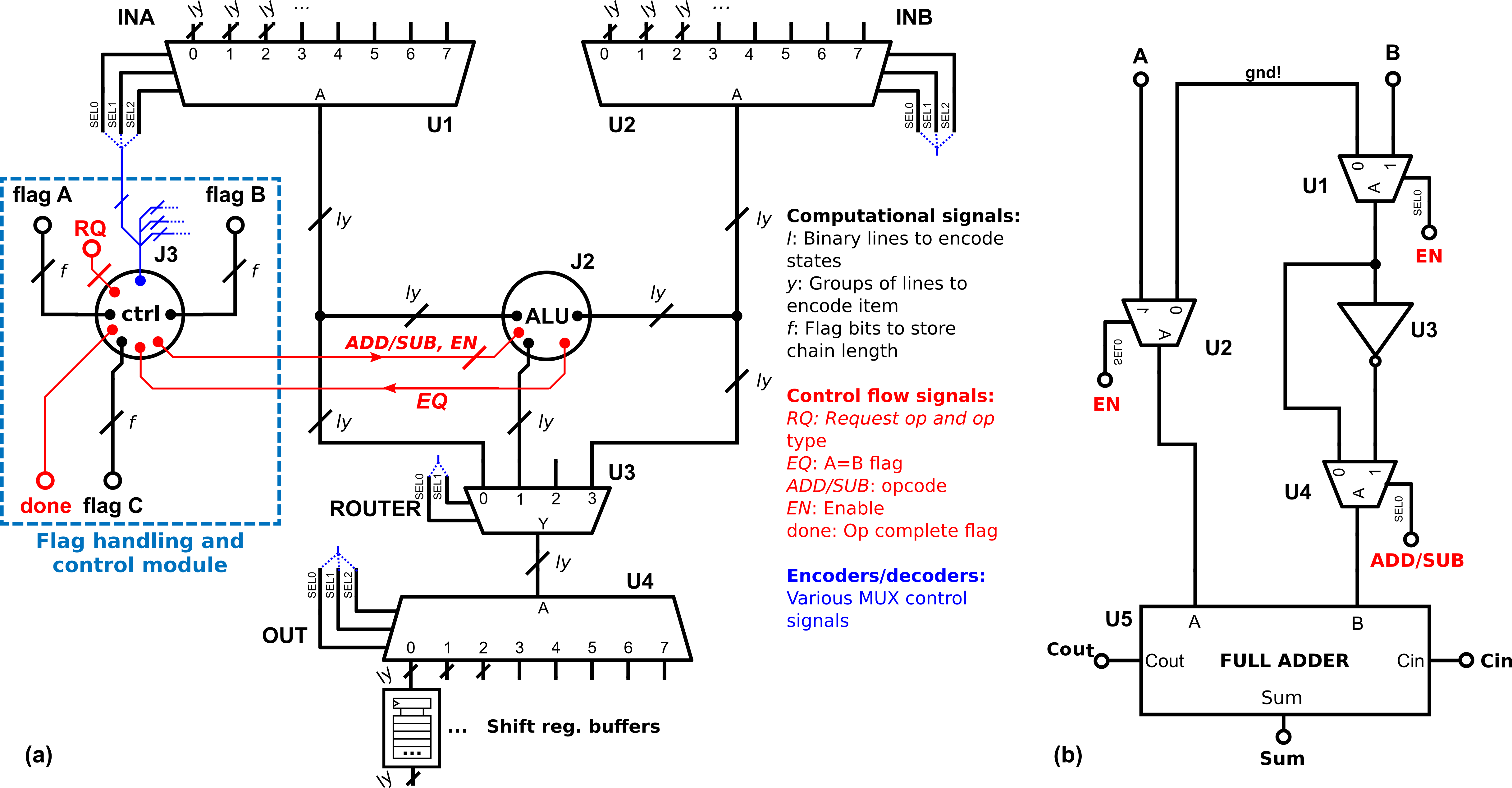}
\caption{Summary of CoPU design. (a) Overview of the entire system. Signal meanings explained within figure. The shift register output buffers are shown at the very bottom of the schematic. (b) Single-bit processing inside the ALU. In the absence of over/underflow issues switching between ADD/SUB modalities simply involves a decision to either bypass an inverter or not.}
\label{sysfig}
\end{figure*}

The proposed holographic representation machinery can be implemented as a fully digital system in a very straightforward manner as shown in the block diagram of Figure \ref{sysfig}. The underlying set will be implicitly determined by the bit-width used. The inverses of each n-vector element under element-wise modular addition are simply their 2's complements. Full representation of any semantic object can therefore consist of $d$, $\log_2p$-bit words, plus $x$ flag bits for tracking the number of items in any given chain.

The superposition operation can be handled by the hardware as `APPEND' operations (akin to linked lists); the system need only know the operands and the state of their flag bits. In practice this would be implemented as $d$ `SELECT' operations, which directly map onto a simple ($l \cdot n$)-width\footnote{$n$ `bundles' of $l$ binary lines.} multiplexer/demultiplexer (MUX/DEMUX) pair. A small digital controller circuit determines the appropriate, successive configurations of the MUX/DEMUX structure depending on the flag bits of the operands (see below). The same circuit also computes and sets the flag bits of the resulting chain. The hardware-level complexity of our proposed system can be contrasted with the standard element-wise addition approach, which requires $n$ times $z = ceil(l)$-level `ADD' operations (cost: $n$, $z$-bit adders, or one time-shared $z$-size adder or valid trade-off solutions in between).

The binding operation can be carried out by $n$ element-wise addition/subtractions (ADD/SUB), implementable as $n$, $z$-bit ADD/SUB modules. Because of the modular arithmetic rules overflow bits are simply ignored. The ADDSUB terminal of each module can directly convert one of the operands into its 2's complement inverse as is standard. This is illustrated in Figure \ref{sysfig}(b). The complexity of (a maximum of) $n$, $z$-bit additions can be contrasted to the computational cost of circular convolution, which would involve $n^2$ multiplication and $n \cdot (n-1)$ additions ($=n \cdot (n-1)$ MACs + $n$ multiplications). On top of this, the additional hardware cost of shifting a chosen operand of the circular convolution $n$ times in its entirety must also be considered.

Finally, the design is completed by a controller unit that orchestrates the operation of the entire system. The unit: i) instructs the arithmetic-logic unit (ALU) what operation to execute (ADD/SUB signal) and when (EN signal), at the behest of a request signal (RQ), ii) is informed by the ALU when the input operands are equal (EQ); useful for e.g. branch-equal-type Assembly-level operations, iii) controls all multiplexers, iv) internally executes the flag arithmetic, and v) outputs an operation termination flag (done). Shift register buffers capture the output of the CoPU and latch it for further use.

Naturally, alternative hardware implementations are also possible. This might include fully analogue ones, e.g. using analogue multiplexers for superposition and current-steering-based binding \cite{Deveugele2006}. Alternatively it might include `packet'-based ones where chains are packaged into e.g. TCP-like (Transmission Control Protocol) packets and communicated across an internet-like router structure. Each packet could contain a header detailing the number of items within the packet and a payload, a technique similar to the protocol used in neuromorphic systems communications over the internet \cite{Boahen2000}. The proposed implementation is chosen because it naturally maps onto easily synthesisable digital hardware. The most efficient implementation technique in any given system, however, will naturally depend on the rest of the system, e.g. on whether the broader environment operates in mainly analogue or digital.

\subsection{CoPU: further details and performance evaluation}

The CoPU from Figure \ref{sysfig} has been designed in Cadence using TSMC's 65nm technology for the purposes of performance evaluation. The CoPU used: $l=8$, $y=1$, $d=8$ (see table \ref{objrel}). Performance was assessed in terms of power efficiency and transistor-count (proxy for area footprint).

\subsubsection{Power performance} The CoPU was assessed for power dissipation when: i) executing an 4-item $\times$ 2-item binding operation, ii) executing an 8-item superposition and iii) in the idle state. In all cases, total system power dissipation figures include: a) the internal power consumption of the system proper, b) the energy spent by minimum-size inverters in order to drive the signal (semantic object) inputs and c) the consumption of the output register buffers. For both superposition and binding, estimated worst case figures are given.

For superposition, worst case is expected to be obtained when transferring the `all elements = 1' (all-1) item into locations where the `all-0' item was previously stored. This is because all bits in both input drivers and output buffers will be flipped by the new input. Furthermore, for our tests the entire system was initialised so that every node started at voltage 0 (GND), which means that the parasitic capacitances from input MUX to output register buffers also needed to be charged to logic 1. In binding, as for superposition, the system is initialised with all inputs (and also outputs) at logic 0. The worst case is expected to be given when adding two all-1 items. This is because all inputs and all outputs bar one need to be changed to logic 1. For example going from the state $0000 + 0000 = 0000$ to $1111 + 1111 = 1110$ requires us to flip all 8 input bits and 3/4 output bits. Additionally we opted for a $4 \times 2$-item binding in order to capture the worst case in handling the flag bits as well (for a binding operation performing a total of eight $1 \times 1$-item suboperations). In both cases a $20\,ns$ clock period ($50\,MHz$) was used and each operation lasted 9 clock cycles.

The performance figures indicate a power breakdown as summarised in table \ref{procpow}. Internal dissipation refers to the power consumed by the system shown in Figure \ref{sysfig}(a), excluding the shift register buffers. Driver dissipation is the consumption of the inverters driving the inputs to the system (not shown in Figure \ref{sysfig}(a)). Register dissipation refers to the buffer registers. Cycles/operation refers to how many clock cycles it takes to conclude the corresponding operation for each full item.

\begin{table}[]
\centering
\caption{CoPU power dissipation performance for worst case corners in both superposition and binding cases. Figures quoted for an 8-item superposition and an $4 \times 2$-item binding. Clock frequency: $20\,ns$.}
\begin{tabular}{l|ccc}
                     & Sup.  & Bind. & Units  \\ \hline
Total energy/op      & 5.97 & 5.79 & pJ     \\ \hline
Internal dissipation & 1.82 & 2.07 & pJ     \\
Driver dissipation   & 0.73 & 0.73 & pJ     \\
Register dissipation & 3.43 & 2.99 & pJ     \\
Cycles/op            & 9    & 9    & -      \\
Time/op              & 180  & 180  & ns     \\
Power @ 50MHz clk    & 33.2 & 32.2 & $\mu$W
\end{tabular}
\label{procpow}
\end{table}

The figures in table \ref{procpow} indicate that most of the power is dissipated in registering the outputs ($>50\%$). Next is the internal power dissipation, most of which occurs in the control module ($\approx$ $1.6-1.7\,pJ$). We further note that superposition and binding cost similar amounts of energy though their internal breakdown is slightly different. The lower buffer register dissipation in binding (we only flip $7/8$ bits at the output in our estimated worst case) is counterbalanced by an increase in energy expenditure for computing the sum of the operands (added internal dissipation). Finally, static power dissipation was calculated at $\approx 82.5\,nW$.

\subsubsection{Transistor count} The transistor count for the overall system and its sub-components is summarised in table \ref{tranct}. We note that the data-path part of the system, which includes the MUX/DEMUX trees and ALU only requires 880 transistors. This means 110 transistors/bit of bit-width, of which 42 in the ALU and 68 in the MUX/DEMUX trees. In larger designs supporting longer item chains the multiplexer tree becomes deeper and adds extra transistors.

\begin{table}[]
\centering
\caption{Transistor count for CoPU and main component parts.}
\begin{tabular}{lr}
Total          & 4382 \\ \hline
Data path      & 880  \\
Control module & 2304 \\
Registers      & 1198
\end{tabular}
\label{tranct}
\end{table}

We conclude with some observations: The CoPU can be constructed using relatively few, simple and standard electronic modules that are all very familiar to the digital designer. The relative costs of both basic operations of superposition and binding are also very similar, in contrast to the large energy imbalance between multiplication and addition carried out using conventional digital arithmetic circuits. Next, we note that the proposed architecture lends itself naturally to speed/complexity trade-offs. First, $2 \cdot d$ DEMUX trees could be implemented in order to allow up to $d$ items to be transferred simultaneously to any location of the output chain. Second, $d$ ALUs could be arrayed in order to perform up to $d \times 1$-item bindings in a single clock cycle. Naturally the increased parallelism would result in bulkier, more power-hungry system versions. Finally, we remark that systems using smaller $l$ in exchange for larger $y$ will in principle be implemented by larger numbers of lower bit-width ALUs operating in parallel. This may simplify the handling of the carry and improve speed (certainly in ripple carry-based designs).

\section{Discussion}\label{disc}

The starting point of this work is the observation that any system consisting of a length $n$ vector with $p$ states per element (corresponding to some fixed number of digital signal lines) can only represent $p^n$ uniquely identifiable vectors. This is effectively a hardware resource constraint and imposes a number of trade-offs warranting design decisions.

Trade-off 1 - expressivity vs. capacity: In the classical holographic representation systems all semantic object vectors are of equal length no matter how many times semantic objects are combined together through superposition or binding. By contrast, in our proposed system some objects will be base items and others will be chains of various lengths. This introduces some constraints into which combinations of semantic objects are allowable, yet the system retains the capability of representing $p^n$ states overall. This seems to be a manifestation of a fundamental trade-off. Cognitive systems may either:

\begin{enumerate}
    \item Operate on relatively few basic semantic objects (objects stored in memory as meaningful/significant) but allow many possible combinations between them, i.e. be expressive but low capacity.
    \item Operate on relatively many basic semantic objects but only accommodate certain possible combinations between them. This is the regime in which our proposed system operates.
\end{enumerate}

We note that the question of the optimum balance between expressivity and capacity is highly complex and requires further study in its own right. In our proposed system capacity and expressivity are to some extent decoupled: $p, y$ affect capacity and expressivity in a trade-off manner whilst $d$ affects only capacity.

Trade-off 2 - `holographicity' vs. compression: Cognitive systems can be conceived at different levels of `holographicity' as determined by the percentage of operations that are operand length-preserving. For fixed maximum semantic object length the choice lies between the extreme of always utilising the full length of $n$ elements in order to represent every possible semantic object (full-holographic), or allowing some semantic objects to be shorter (non-holographic). This significantly impacts the amount of information each numerical element carries. In a fully holographic representation transmitting or processing even a single-item-equivalent semantic object requires handling of $n$ elements; the same as transmitting/processing the equivalent of a long chain. The semantic information per element may dramatically differ in each situation. In our proposed system, however, superpositions of fewer items are represented by shorter chains. This illustrates how less holographic systems generally offer the option of operating on more compressed information, i.e. closer to the signal-to-noise ratio (SNR) limit.

Naturally there is a price to pay for compression: when creating new semantic objects for storage it is extremely useful if these new objects can be mapped onto minimum-length units (the semantic object basis of any cognitive system). Mechanisms for mapping any arbitrary chain onto such units need to be supported, adding to system complexity. Furthermore, in a non-holographic system any circuitry designed to support the last items of a chain may be utilised only infrequently. This is expected to strongly affect hardware design decisions. %In terms of energy efficiency we point out the potential connectivity costs in terms of parasitic capacitances. For example if we have a semi-holographic system where each numerical element of an item can be routed to any one out of $d$ possible destinations as a result of e.g. superposition we need a $log_2d$-depth MUX/DEMUX tree pair (see Figure XXX). If the system were completely non-holographic and any element in any operand could be routed to any element in the result, then a $log_2n$-depth MUX/DEMUX tree pair would be necessary, with all the parasitic capacitance this entails.

%Finally, the trade-off is further complicated when considering the rules of semantic object n-vector lengthening. Systems may be designed to lengthen a semantic object hypo- or hyper-linearly\footnote{Or indeed just linearly.}. The extent to which this is possible to compensate against by allowing shorter unit semantic object lengths is not obvious. This requires dedicated further study.

Trade-off 3 - long vectors with few states per element vs short vectors with many states per element: If we have a fixed number of binary lines (i.e. $l \cdot y = C$), we have a choice of treating $C$ as either: i) one single, large identifier number, ii) a collection of binary bits independent of one another or iii) certain possibilities in between. For example, for $C=16$ we can have $\{l,y\} \in \{(1,16), (2,8), (4,4), (8,2), (16,1)\}$. The number of states we can represent remains fixed at $2^{ly}$, but:

\begin{itemize}
    \item The distance relationships between semantic objects will be different in each case. In the case (1,16) our item consists of a vector of 16x 1-bit elements, and therefore there are 16 nearest neighbours each item (all items that differ from the base object at exactly one position). In the case (16,1) our item is a single 16-bit number which has exactly two nearest neighbours (the elements/items different from the base object by one unit of distance). Note that the case (1,16) corresponds tightly to the spatter code system proposed by Kanerva \cite{Kanerva1997} since modular addition now reduces to a simple XOR.
    \item The degree of modularity achievable in hardware may be impacted in each case. The (1,16) case requires 16x XOR gates in order to perform one item-item binding whilst in the (16,1) case requires a single 16-bit adder. In the case of large values of $C$ there may be an additional impact on speed (how viable is to make a 512-bit adder that computes an answer in one clock cycle/step? - 512x XOR gates on the other hand will compute 512 outputs in one step). This subject requires further, dedicated study.
\end{itemize}

Trade-off 4 - operation complexity vs. property attractiveness: As a rule of thumb operations with more attractive mathematical properties tend to introduce computational and implementational difficulties. This is perhaps well exemplified by examining different binding operations:

\begin{itemize}
    \item Convolution commutes, `scrambles' the information well\footnote{The result bears in general very little resemblance to either of the operands.} and preserves information. However, it lengthens the vectors that it processes and it is computationally heavy (many MACs).
    \item Circular convolution commutes and scrambles. Lengthening no longer occurs, but information is lost and the operation is still heavy on MACs.
    \item Modular arithmetic commutes. Lengthening does not occur and the operation is MAC-lightweight, but information is lost and the scrambling properties are similar to those of superposition by element-wise addition, so the similarity requirements for defining two semantic objects as corrupted versions of each other have to be substantially tightened.
\end{itemize}

Ultimately, a complex mix of factors/specs in all trade-off directions will determine the best cognitive system implementation. This may depend on the overall cognitive capabilities required of the system. In this work we have focussed on a partially holographic system based on effectively multiplexing and addition as the system operations. The advantage of this implementation vs. the holographic approach that we have used as standard and inspiration is that both operations have been simplified in hardware: superposition became a multiplexing operation instead of addition whilst binding became element-wise addition instead of circular convolution. The balance of these advantages vs. the attributes that had to be traded-away (mathematical elegance, full holographicity, etc.) needs to be considered very carefully. In general, however the system is designed for occasions where we have partially restricted expressivity (notable cap on chain length - effective number of successive superpositions allowed) but enables extreme implementational simplicity and high energy efficiency.

Finally, we envision that our proposed CoPU will form a core component of larger systems with cognitive capability. Much like in a traditional computer, our CPU-equivalent will need a memory to which it can communicate as well as peripheral structures. Work in that general direction has very recently begun to gain traction \cite{Rahimi2018, Graves2014}. Relating this back to biological brains we see the closest analogue of our CoPU in the putative attentional systems of the brain; the contents of the input buffers at any given time could be interpreted as the semantic objects in the machine's `conscious attention'. In conclusion, we envisage that future thinking machines will be complex systems consisting of multiple, heterogeneous modules including ANNs, memories (bio-inspired or standard digital look-up tables), sensors, possibly even classical microprocessors and more; all working together to give rise to cognitive intelligence. We hope that our CoPU will play a central role in this `hyperarchitecture' structure by acting as the equivalent of the CPU in a classical computer, and that it will do so with the energy efficiency required for enabling widespread adaptation of cognitive computers.

\section*{Acknowledgements}

The authors would like to thank Prof. Chris Eliasmith whose work provided much of the inspiration for this work.
We also thank Prof. Jesse Hoey for his support and fruitful discussions.

\bibliographystyle{ieeetran}
\bibliography{HRlib}

\begin{IEEEbiography}
[{\includegraphics[width=1in,height=1.25in,clip,keepaspectratio]{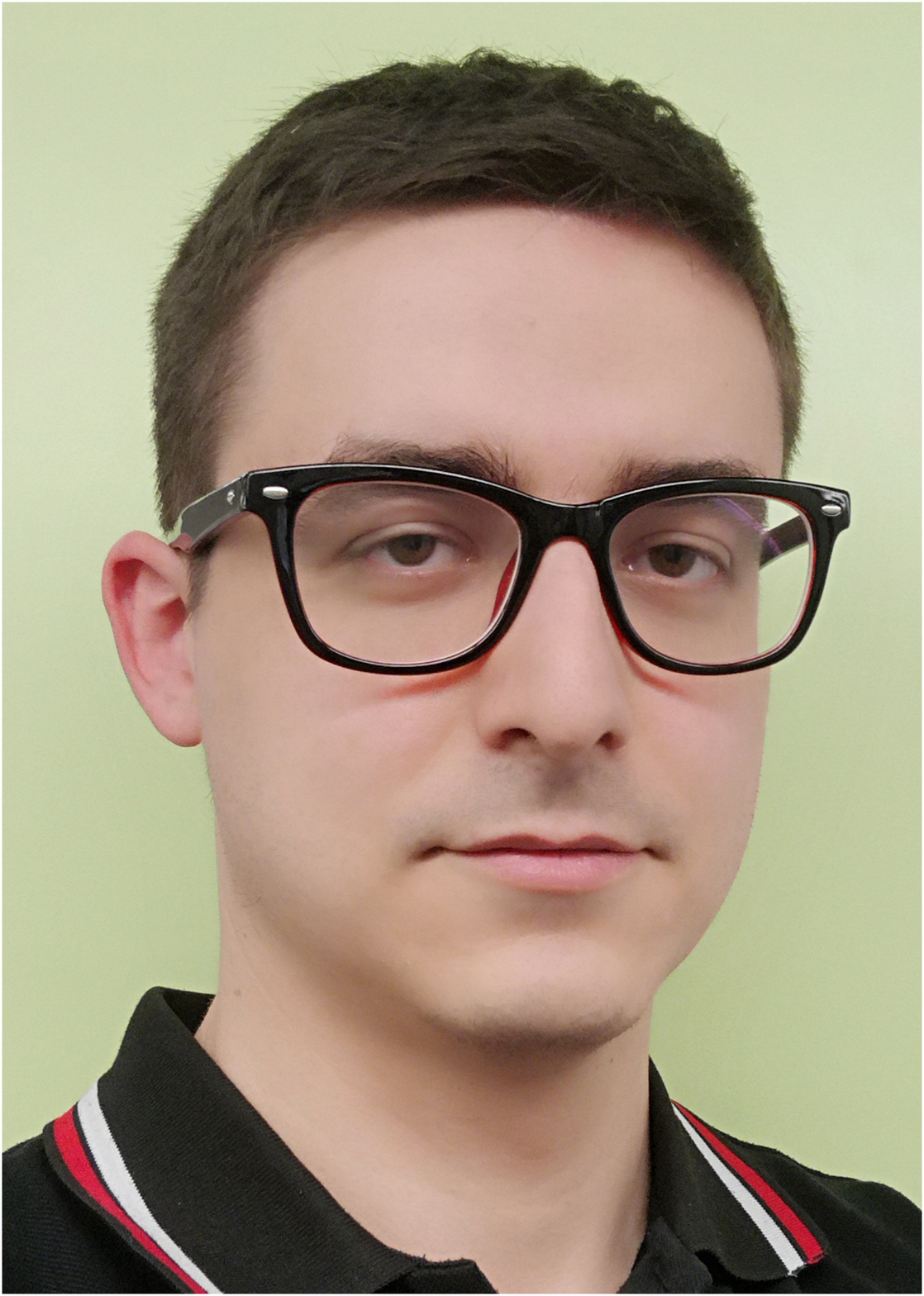}}]{Alexantrou Serb} (M ‘11) received his degree in Biomedical Engineering from Imperial College in 2009 and his PhD in Electrical and Electronics Engineering from Imperial College in 2013. Currently he is a research fellow at the Zepler Institute (ZI) dept., University of Southampton, UK. His research interests are: cognitive  computing, neuro-inspired engineering, algorithms and applications using RRAM, RRAM device modelling and instrumentation design.
\end{IEEEbiography}

\begin{IEEEbiography}
[{\includegraphics[width=1in,height=1.25in,clip,keepaspectratio]{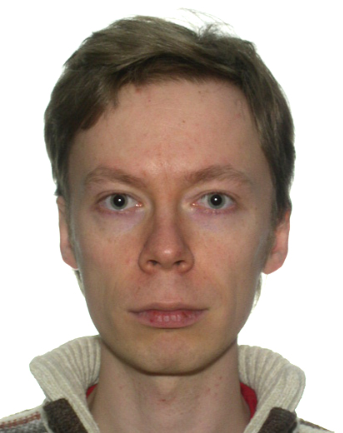}}]{Ivan Kobyzev} received his Master's degree in Mathematical Physics from St Petersburg State University, Russia, in 2011 and his PhD in Mathematics from Western University, Canada, in 2016. While working on the paper he was a postdoctoral fellow at the school of computer science, University of Waterloo, Canada. His research interests are: artificial intelligence, machine learning, cognitive  computing, representation learning.
\end{IEEEbiography}

\begin{IEEEbiography}
[{\includegraphics[width=1in,height=1.25in,clip,keepaspectratio]{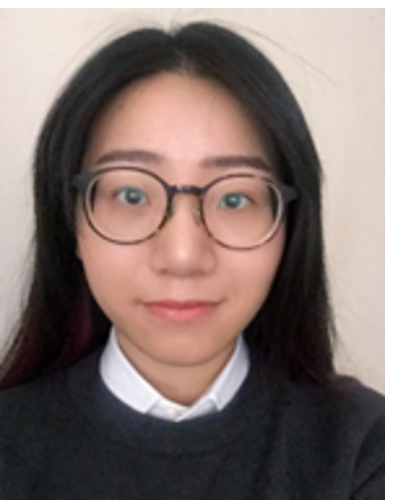}}]{Jiaqi Wang} received her bachelor degree in Microelectronic Science and Engineering from Shenzhen University, China, in 2017 and her M.Sc. degree in Microelectronics Systems Design from University of Southampton, UK, in 2018. And she is currently pursuing her PhD studies in University of Southampton, working towards the memristor-based hardware design, integrated circuit designs and cognitive computing. 

\end{IEEEbiography}

\begin{IEEEbiography}
[{\includegraphics[width=1in,height=1.25in,clip,keepaspectratio]{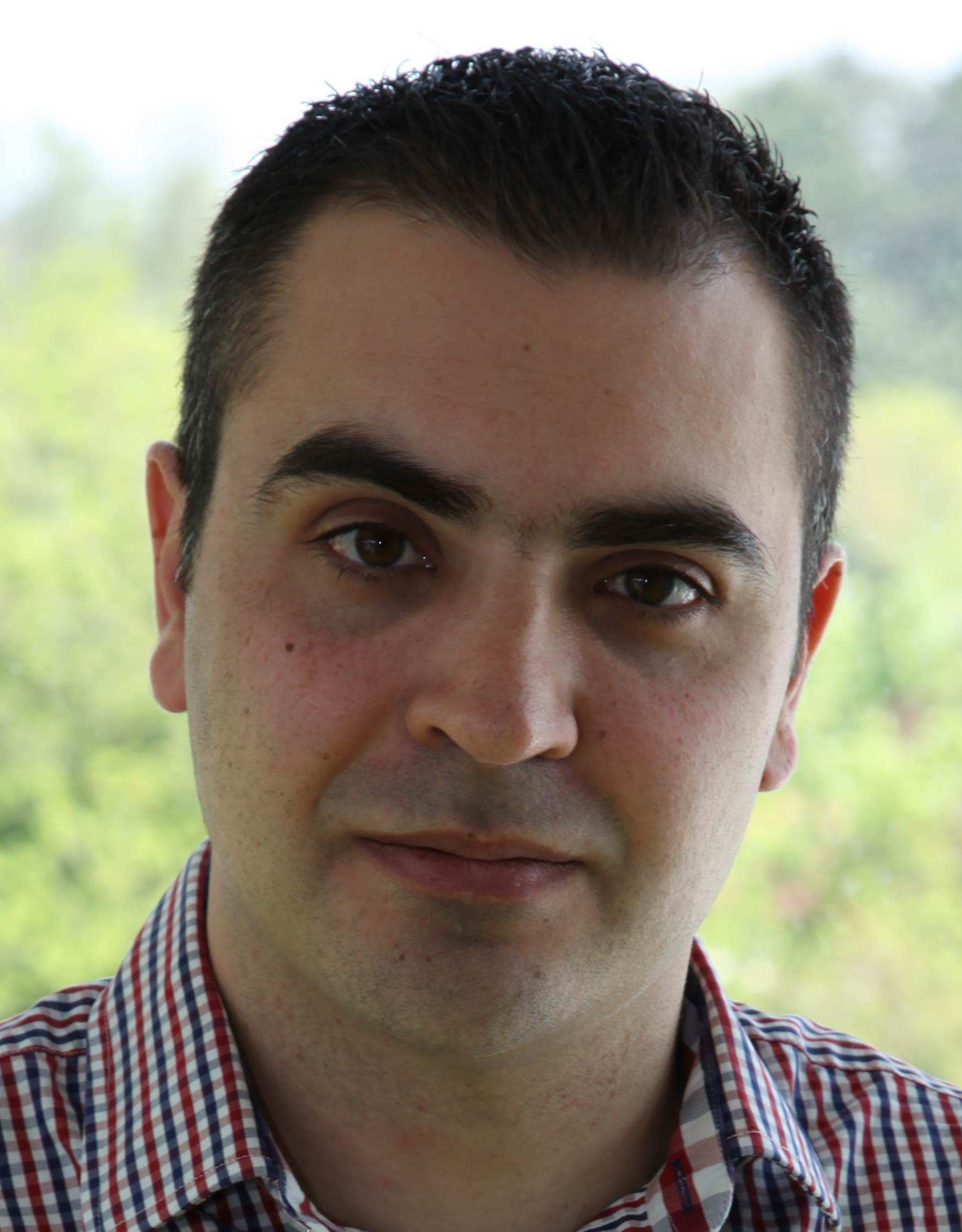}}]{Themis Prodromakis} (SM’08) is Professor of Nanotechnology and Head of the Electronic Materials and Devices Research Group in the Zepler Institute for Photonics and Nanoelectronics at the University of Southampton. His background is in Electron Devices and nanofabrication techniques, with his research being focused on bio-inspired devices for advanced computing architectures and biomedical applications.
\end{IEEEbiography}

\vfill

\end{document}